\documentclass[10pt,twocolumn,letterpaper]{article}

\usepackage{cvpr}
\usepackage{times}
\usepackage{epsfig}
\usepackage{graphicx}
\usepackage{subfigure}
\usepackage{amsmath}
\usepackage{amssymb}

\cvprfinalcopy

\ifcvprfinal\fi
\begin{document}

\title{Cross-View Image Matching for Geo-localization in Urban Environments}

\author{Yicong Tian, Chen Chen, Mubarak Shah\\
Center for Research in Computer Vision (CRCV), University of Central Florida (UCF)\\
{\tt\small tyc.cong@gmail.com, chenchen870713@gmail.com, shah@crcv.ucf.edu}
}

\maketitle

\begin{abstract}
   In this paper, we address the problem of cross-view image geo-localization. Specifically, we aim to estimate the GPS location of a query street view image by finding the matching images in a reference database of geo-tagged bird's eye view images, or vice versa.
   To this end, we present a new framework for cross-view image geo-localization by taking advantage of the tremendous success of deep convolutional neural networks (CNNs) in image  classification and object detection. First, we employ the Faster R-CNN \cite{ren2015} to detect buildings in the query and reference images. Next, for each building in the query image, we retrieve the $k$ nearest neighbors from the reference buildings using a Siamese network trained on both positive matching image pairs and negative pairs. To find the correct NN for each query building, we develop an efficient multiple nearest neighbors matching method based on dominant sets.
   We evaluate the proposed framework on a new dataset that consists of pairs of street view and bird's eye view images. Experimental results show that the proposed method achieves better geo-localization accuracy than other approaches and is able to generalize to images at unseen locations.
\end{abstract}

\section{Introduction}
Geo-localization is the problem of determining the real-world geographic location (\eg GPS coordinates) of {\em each pixel} of a query image. It plays a key role in a wide range of real-world applications such as target tracking, change monitoring, navigation, etc. Traditional geo-localization approaches deal with satellite and aerial imagery that usually involve different image sensing platforms and require accurate sensor modeling and  pixel-wise geo-reference image, \eg digital ortho-quad (DOQ), \cite{zitova2003} and Digital Elevation Map (DEM). Recently, image geo-localization methods have been devised for coarse image level geo-localization instead of pixel-wise geo-localization pursued in traditional geo-localization methods. In particular, this problem has attracted considerable attention due to the availability of ground-level geo-tagged imagery \cite{hays2008,zamir2010,torii2011,zamir2014,Shan2014,schindler2007}.
In these methods, the geo-location of a query image is obtained by finding its matching
reference images from the same view (\eg ground-level Google Street View images), based on the assumption that a reference dataset consisting of geo-tagged images is available.
However, such geo-tagged reference data may not be available. For example, ground-level images of some geo-graphical locations do not have geo-location information.

\begin{figure}[t]
\begin{center}
   \includegraphics[width=0.95\linewidth]{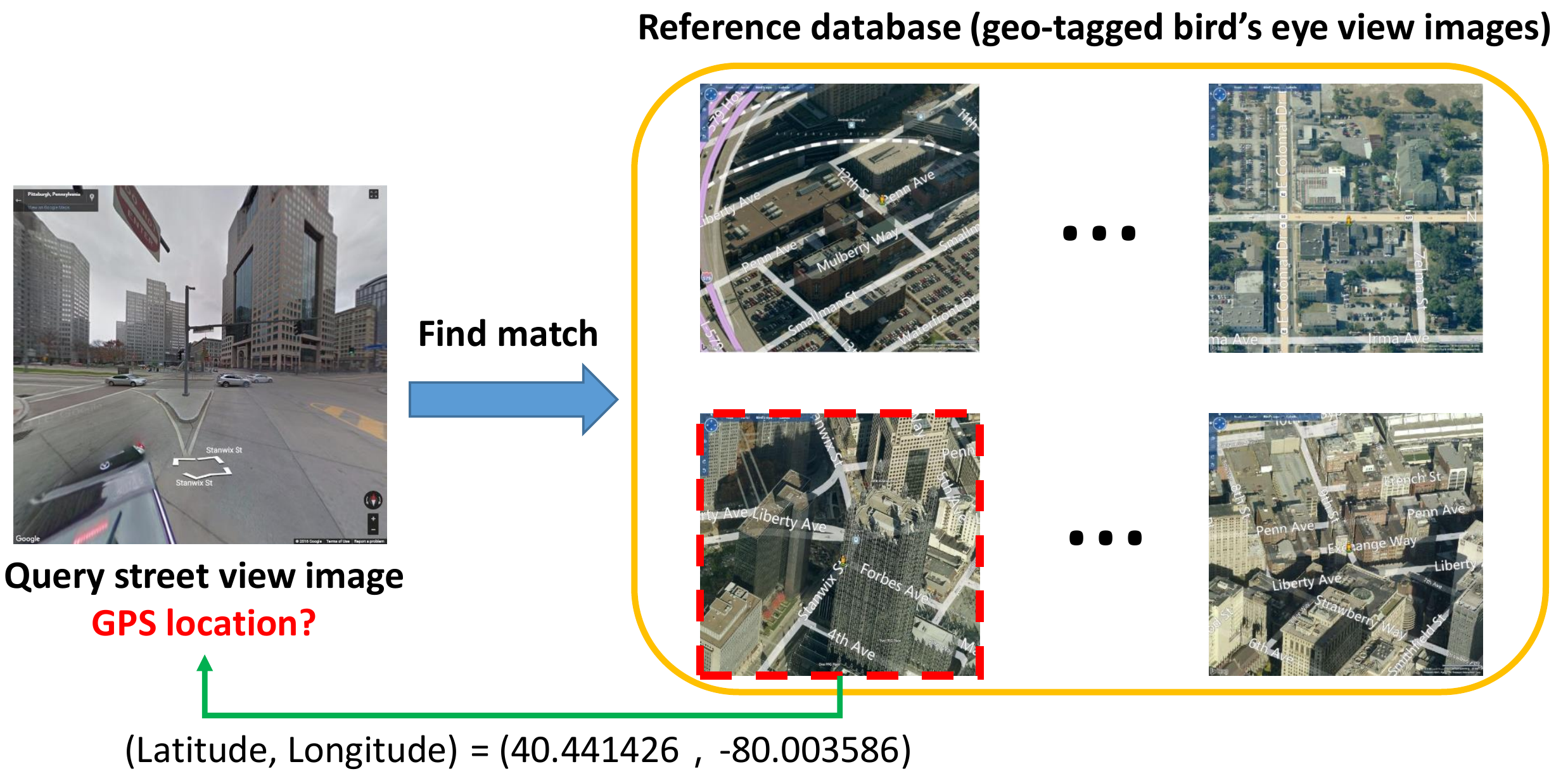}
\end{center}
   \caption{An example of geo-localization by cross-view image matching. The GPS location of a street view image is predicted by finding its match in a database of geo-tagged bird's eye view images.}
\label{fig:cross_match}
\end{figure}

\begin{figure}[t]
\begin{center}
   \includegraphics[width=0.95\linewidth]{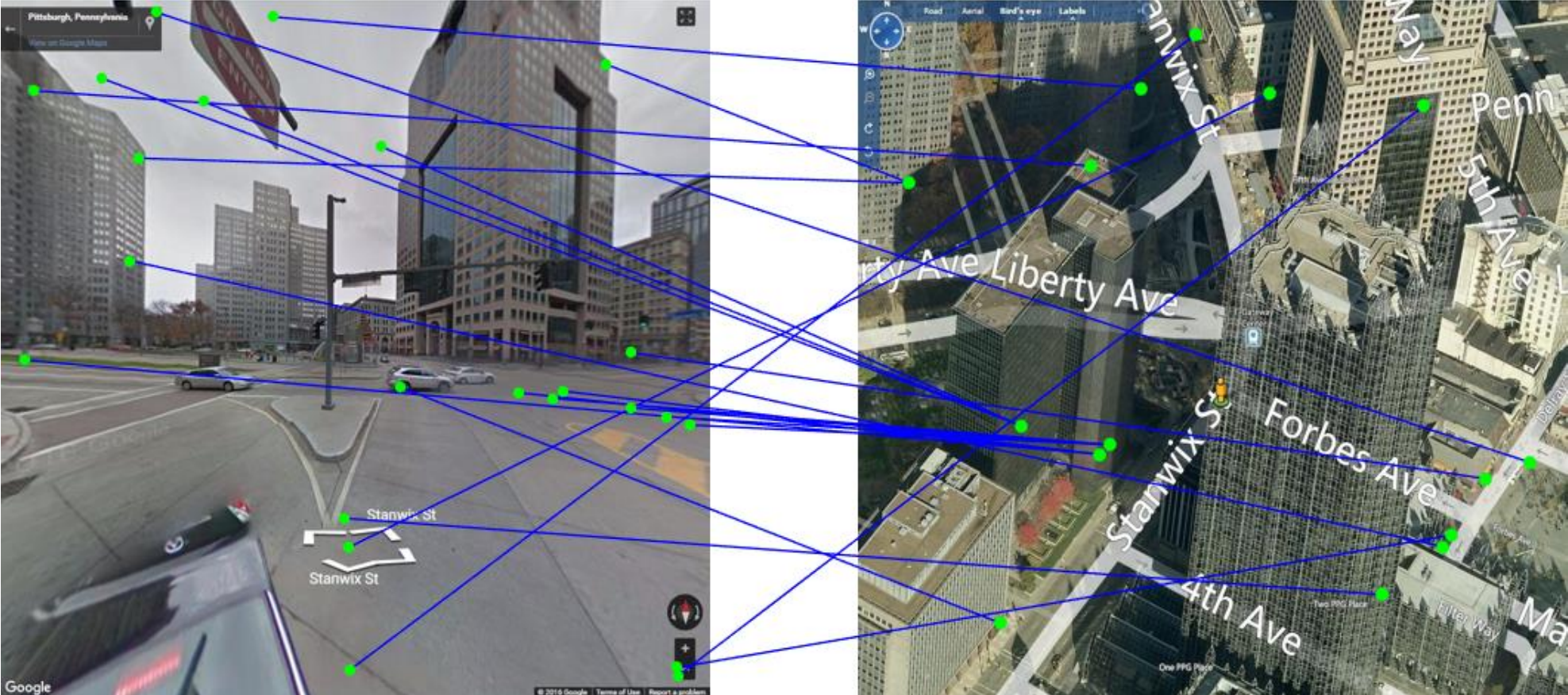}
\end{center}
   \caption{SIFT points matching between two cross-view images. The matching fails due to very different visual appearance under different viewpoints.}
\label{fig:sift_match}
\end{figure}

An alternative is to predict the geo-location of a query image by finding its matching
reference images from some other views. For example, predict the geo-location of a query street view image based on a reference database of bird's eye view images (see Figure~\ref{fig:cross_match}).
This becomes a cross-view image matching problem, which is very challenging because of the following reasons. 1) Images taken from different viewpoints are visually different. 2) The images may be captured with different lighting conditions and during different seasons. 3) The mapping from one viewpoint to the other may be highly non-linear and very complex. 4) Traditional low-level features like SIFT, HOG, etc. may be very different for cross-view images as shown in Figure~\ref{fig:sift_match}.

Historically, viewpoint invariance has been an active area of research in computer vision. Some of this work was inspired by classic work of Biederman on recognition-by-components theory \cite{biederman1987recognition}. It explains how humans are able to recognize objects by separating them into geons, which are based on 3D-shape like cylinders and cones. One important factor of this theory is view-invariance properties of edges \ie curvature, parallel lines, co-termination, symmetry and co-linearity. In computer vision, over the years it has been demonstrated that directly detecting 3D shapes from 2D images is a very difficult problem. However, some of the view-invariance properties \eg scale and affine invariance have been successfully used in local descriptors work of Lowe \cite{lowe2004distinctive} and Mikolajczyk \cite{scale2004affine}. However, as illustrated in Figure~\ref{fig:sift_match}, SIFT point matching in high oblique view fails. In this paper, we investigate deep learning approaches for this problem and present a new cross-view image matching framework for geo-localization
 by automatically detecting, representing and matching the semantic information in cross-view images. Instead of matching local features \eg SIFT and HOG, we perform cross-view matching based on buildings, which are semantically more meaningful and robust to viewpoints. Therefore, we first employ the Faster R-CNN \cite{ren2015} to detect buildings in the query and reference images. Then, for each building in the query image, we retrieve the $k$ matching nearest neighbors (NNs) from the reference buildings using a Siamese network \cite{chopra2005} trained on both positive and negative matching image pairs. The network learns a feature representation that transfers the original cross-view images to a lower dimensional feature space. In this learned feature space, matching image pairs are close to each other and unmatched image pairs are far apart.
 To predict the geo-location of the query image, taking the location of the first nearest neighbor in reference images may not be optimal because in most cases the first nearest neighbor does not correspond to the correct match. Since the GPS locations of the detected buildings in the query image is close, the GPS locations of their matched buildings in reference images should be close as well. Therefore, besides local matching (matching individual buildings), we also enforce a global consistency constraint in our geo-localization approach.
 To solve this problem instead of relying on the first nearest neighbor, we employ multiple nearest neighbors and develop an efficient multiple nearest neighbors matching method based on dominant sets \cite{pavan2007}. The nodes in dominant sets form a coherent and compact set in terms of pairwise similarities. The final geo-localization result is obtained by taking the mean GPS location of the selected reference buildings in the dominant set.

The main contributions of this paper are three-fold:

$\bullet$ We present a new image geo-localization framework by matching a query street view (or bird's eye view) image to a database of geo-tagged bird's eye view (or street view) images. In contrast to the existing works, which either match street-view imagery with street-view imagery or street view queries to aerial imagery, we consider both directions to comprehensively evaluate our approach.

$\bullet$ We develop an efficient multiple nearest neighbors matching
method based on dominant sets, which is fast and scalable to large scale.

$\bullet$ We introduce a new large scale dataset which consists of pairs of annotated street view and bird's eye view images collected from three different cities in the United States.


\section{Related Work}
\label{sec:related_work}

\subsection{Ground-level Geo-localization}
The large collections of geo-tagged images on the Internet have fostered the research in geo-localization using ground-level imagery \eg street view images \cite{schindler2007,hays2008,zamir2010,torii2011,zamir2014,Tobias2016}.
One assumption is that there is a reference dataset consisting of geo-tagged images. Then, the problem of geo-locating a query image boils down to image retrieval.
The geo-locations of the matching references are utilized to determine the location of the query image.

Schindler \etal~\cite{schindler2007} explored geo-informative features on
specific locations of a city to build vocabulary trees for city-scale geo-localization.
Hays and Efros \cite{hays2008} proposed the IM2GPS method to characterize geo-graphical information of query images as probability distributions over the
Earth's surface by leveraging millions of GPS-tagged images.
Zamir and Shah \cite{zamir2014} extracted both local and global appearance features from images and employed the Generalized Minimum Clique Problem
(GMCP) \cite{feremans2003} for features matching between query and reference street-view images.
Recently, Weyand \cite{Tobias2016} introduced PlaNet, a deep learning model that integrates several cues from images, for photo geo-localization and demonstrated superior performance over IM2GPS \cite{hays2008}.

\subsection{Cross-view Geo-localization}
Although ground-level image-to-image matching approaches have achieved promising results, however, due to the fact that only small number of cities in the world are covered by ground-level imagery, it has not been feasible to scale up  this approach to  global level.
On the other hand, a more complete coverage for overhead reference data such as satellite/areial
imagery and digital elevation model (DEM) has spurred a growing interest in cross-view geo-localization \cite{bansal2012ultra,lin2013,bansal2011geo,workman2015,lin2015,ozcanli2016,hays2016eccv}.

\begin{figure*}[t]
\begin{center}
   \includegraphics[width=0.9\linewidth]{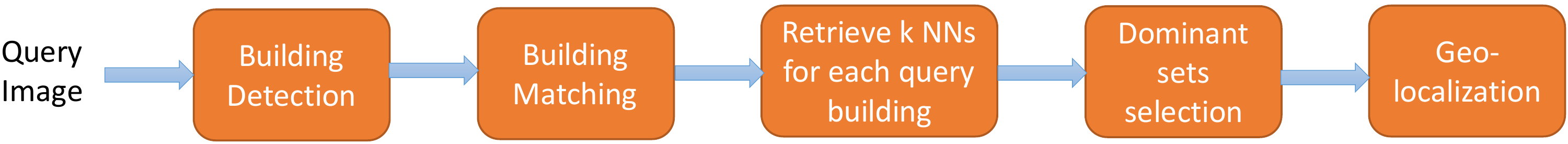}
\end{center}
   \caption{The pipeline of the proposed cross-view geo-localization method.}
\label{fig:pipeline}
\end{figure*}

Lin \etal~\cite{lin2013} proposed a cross-view image geo-localization approach using a training triplet including query ground-level images, the corresponding
reference aerial images and land cover attribute maps to learn the feature translation between cross-view images.
Bansal \etal~\cite{bansal2012ultra} developed a method for matching facade imagery from
different viewpoints relying on the structure of self-similarity of patterns on facades. A scale-selective self-similarity descriptor was proposed for facade extraction and segmentation. Given all labeled descriptors in the bird's eye view database, facade matching of the street-view
queries was done in a Bayesian classification framework.
Lin \etal~\cite{lin2015} investigated the deep learning method for cross-view image geo-localization.
A deep Siamese network \cite{chopra2005} was used to learn feature embedding for image matching. One important limitation of this method is that it requires scale and depth meta data for street-view query images during testing,which is unrealistic.
Workman \cite{workman2015} used existing CNNs to transfer ground-level image feature representation to aerial images via a cross-view training procedure.
Vo \etal~\cite{hays2016eccv} explored several CNN architectures with a new distance based logistic loss for matching ground-level query images to overhead satellite images. Rotational invariance and orientation regression were incorporated during training to improve geo-localization accuracy.

In general, our method differs from the existing cross-view image matching approaches in three main aspects:

$\bullet$ We propose to use buildings as the reference objects to perform image matching. Such semantic information is more meaningful and robust to changes in viewpoint than local appearance-based features.

$\bullet$ We perform geo-localization by multiple nearest neighbors matching. Moreover, unlike the existing cross-view image matching approaches which find the corresponding match reference images for the (single) query image, our method extends to multiple queries (\ie buildings in a query image) matching, which provides a more flexible and accurate solution by taking the global consistency into account.

$\bullet$ Finally, we do not require depth map or other meta data in our approach.

\section{Proposed Cross-view Geo-localization Method}
\label{sec:method}

The pipeline of the proposed method for cross-view geo-localization is shown in Figure~\ref{fig:pipeline}. In the following subsections, we describe each step of our approach.

\subsection{Building Detection}

 To find the matching image or images in the reference database for a query image, we resort to match buildings between cross-view images since the semantic information of images is more robust to viewpoint variations than appearance features. Therefore, the first step is to detect buildings in images. We employ the Faster R-CNN \cite{ren2015} to achieve this goal due to its state-of-the-art performance for object detection and real-time execution.  Faster R-CNN effectively unifies the convolutional
region proposal network (RPN) with the Fast R-CNN \cite{girshick2015} detection network by sharing image convolutional features. The RPN is trained end to end in an alternating fashion with the Fast R-CNN network to generate high-quality region proposals.
In our application, the detected buildings in a query image serve as query buildings for retrieving the matching buildings in the reference images.

\subsection{Building Matching}
For a query building detected from the previous building detection phase, the next step is to search for its matches in the reference images with known geo-locations. Our goal is to find a good feature representation for cross-view images so that we can accurately retrieve the matched reference images for a query image.

The Siamese network \cite{chopra2005} has been utilized in image matching \cite{lin2015,zagoruyko2015}, tracking \cite{tao2016} and retrieval \cite{wang2015}.
We adopt this network structure to learn deep representations in order to distinguish matched and unmatched building pairs in cross-view images. Let $X$ and $Y$ denote the street view and  bird's eye view image training sets respectively. A pair of building images $x \in X$ and $y \in Y$ are used as input to the Siamese network which consists of two deep CNNs sharing the same architecture. $x$ and $y$ can be a matched pair or a unmatched pair. The objective is to automatically learn a feature representation, $f(\cdot)$, that effectively maps $x$ and $y$ from two different
views to a feature space, in which matched image pairs are close to each other and unmatched image pairs are far apart.
In order to train the network towards this goal, the Euclidean distance of the matched pairs in the feature space should be small (close to 0) while the distance of the unmatched pairs should be large. We employ the contrastive loss \cite{hadsell2006}:
\begin{equation}
L(x,y,l) = \frac{1}{2} lD^2 + \frac{1}{2}(1-l) \left\{ \operatorname{max} (0, m-D) \right\}^2,
\label{eq:contrastive_loss}
\end{equation}
where $l \in \{0,1\}$ indicates if $x$ and $y$ is a matched pair, $D$ is the Euclidean distance between the two feature vectors $f(x)$ and $f(y)$, and $m$ is the margin parameter.
%

\subsection{Geo-localization Using Dominant Sets}

A simple approach for geo-localization will be, for each detected building in the query image, take the GPS location of its nearest neighbor in reference images, according to building matching. However, this will not be optimal. In fact, in most cases the nearest neighbor does not correspond to the correct match. Therefore, besides local matching (matching individual buildings), we introduce a global constraint to help make better geo-localization decision. In a given query image, typically there are multiple buildings and their GPS locations should be close. Therefore, the GPS locations of their matched buildings should be close as well. This is our global constraint during geo-localization.

\begin{figure*}[t]
\begin{center}
   \includegraphics[width=\linewidth]{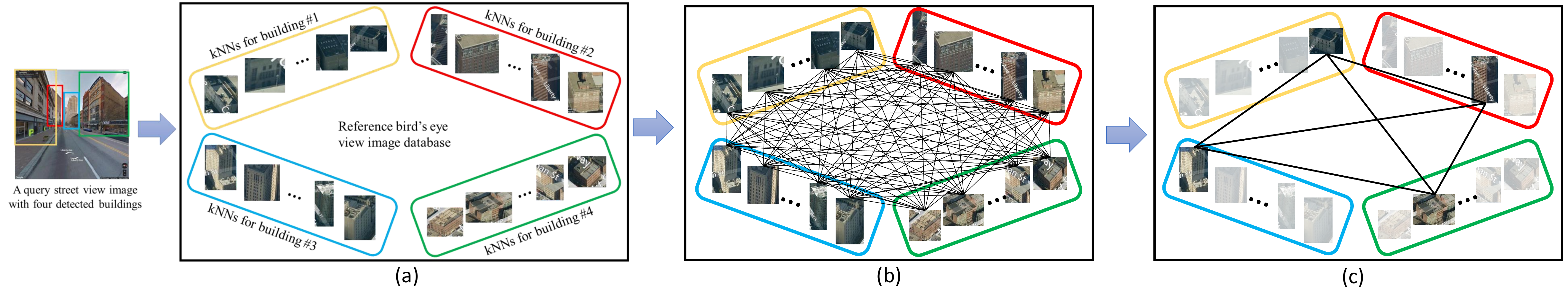}
\end{center}
   \caption{An example of geo-localization using dominant sets. Given a query street view image (shown on the left) with four detected buildings, a cluster is formed for each query building by taking its $k$ nearest neighbors in reference images (a). A graph is built using all the selected reference buildings (b). Dominant sets algorithm is applied to select the best set of reference buildings both in terms of global and local similarities (c). The final geo-localization result is obtained by taking the mean GPS location of the four selected reference buildings in dominant set.}
\label{fig:geolocalization}
\end{figure*}

For each detected building in the query image, $k$ nearest neighbors are selected from reference images based on building matching scores. The nearest neighbors for each query building form a cluster as shown in Figure \ref{fig:geolocalization}. An undirected edge-weighted graph $G=(V,E)$ with no self-loops is built using all the selected reference buildings. Here, $V=\{1,\dots,n\}$ represents the set of nodes, one for each selected reference building. $E$ represents the edges. Every pair of nodes which are not in the same cluster are connected by an edge. A weight is associated with each edge, reflecting similarity between pairs of linked nodes. Let the graph $G$ be represented by an $n\times n$ non-negative symmetric matrix $A=a_{ij}$, where elements of this matrix are populated by
\begin{equation}
a_{ij}=\left\{
\begin{aligned}
& \frac{1}{2}(e^{-\frac{d_{ij}^2}{2\sigma ^2}}+\alpha(s_i+s_j)) & \quad \text{if } (i,j)\in E,\\
& 0 & \quad \text{otherwise.}
\end{aligned}
\right.
\end{equation}
When node $i$ and node $j$ are connected by an edge, $a_{ij}$ denotes the edge weight which measures the similarity between reference buildings $i$ and $j$. $d_{ij}^2$ is the distance between $i$ and $j$'s GPS locations (obtained from their corresponding images) in Cartesian coordinates, which is a global measure. $s_i$ is the similarity between query building and reference building $i$ based on their building matching score, which is a local measure. Therefore edge weights incorporate both local matching information and GPS-based global constraint. The goal of geo-localization is to select at most one reference building from each of the cluster, such that the total weight is maximized.

We use dominant sets \cite{pavan2003new,pavan2007} to solve this problem. For a non-empty subset $S\subseteq V$, $i\in S$ and $j\notin S$, define
\begin{align}
\phi_S(i,j)=a_{ij}-\frac{1}{\lvert S\rvert}\sum_{k\in S}a_{ik},
\end{align}
which measures the relative similarity between nodes $i$ and $j$, with respect to the average similarity between node $i$ and its neighbors in $S$. Then a weight defined recursively as following is assigned to each node $i\in S$:
\begin{equation}
w_S(i)=\left\{
\begin{aligned}
& 1 & \quad \text{if } \lvert S\rvert =1,\\
& \sum_{j\in S\setminus \{i\}}\phi_{S\setminus \{i\}}(j,i)w_{S\setminus \{i\}}(j) & \quad \text{otherwise.}
\end{aligned}
\right.
\end{equation}
$w_S(i)$ measures the overall similarity between node $i$ and the nodes of $S\setminus \{i\}$, with respect to the overall similarity among the nodes in $S\setminus \{i\}$. If $w_S(i)$ is positive, adding node $i$ into its neighbors in $S$ will increase the internal coherence of the set. On the contrary, if
$w_S(i)$ is negative, the internal coherence of the set will be decreased if $i$ is added to its neighbor. Finally, the total weight of $S$ is defined as
\begin{align}
W(S)=\sum_{i\in S}w_S(i).
\end{align}
A non-empty subset of nodes $S\subseteq V$ such that $W(T)>0$ for any non-empty $T\subseteq S$, is said to be a dominant set if
\begin{itemize}
  \item $w_S(i)>0$, for all $i\in S$.
  \item $w_{S\bigcup i}(i)<0$, for all $i\notin S$.
\end{itemize}

We use replicator dynamics algorithm to select a dominant set \cite{pavan2003new,pavan2007}. The nodes in a dominant set form a coherent set both in terms of global and local measures. The final geo-localization result is obtained by taking the mean GPS location of selected reference buildings in the dominant set.

\begin{figure}[htbp]
\begin{center}
   \includegraphics[width=0.95\linewidth]{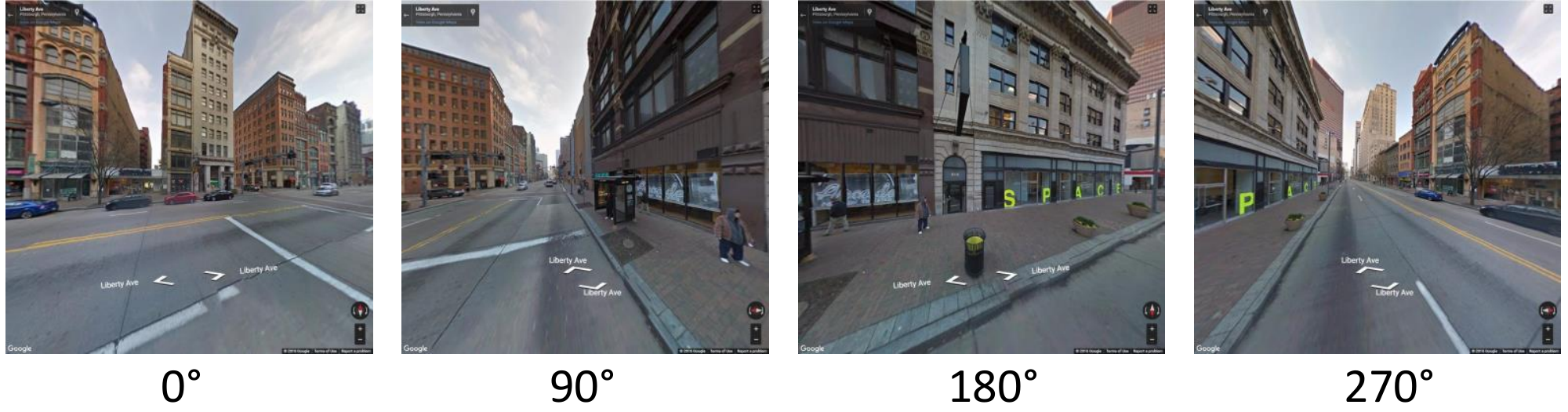}
\end{center}
   \caption{Example street view images with four different camera heading directions at the same GPS location.}
\label{fig:4views}
\end{figure}

In our dataset, four street view images and four bird's eye view images are taken at each GPS location. Each set of four images correspond to camera heading directions of $0^\circ$, $90^\circ$, $180^\circ$ and $270^\circ$. When only one image is used as query, the number of query buildings is usually small. Typically, $4$ query buildings are used for multiple nearest neighbors matching in Figure \ref{fig:geolocalization}. To improve geo-localization accuracy, we propose to use a set of four images with different camera heading directions as query. Figure \ref{fig:4views} shows an example set of street view images with different heading directions. When they are used as query, more query buildings ($12$ in this example) are detected and used for matching, thus improving the geo-localization accuracy.

\section{Experiments}
\label{sec:experiment}

\subsection{Dataset}

\begin{figure*}[t]
\centering
\subfigure{
\includegraphics[width=0.3\linewidth]{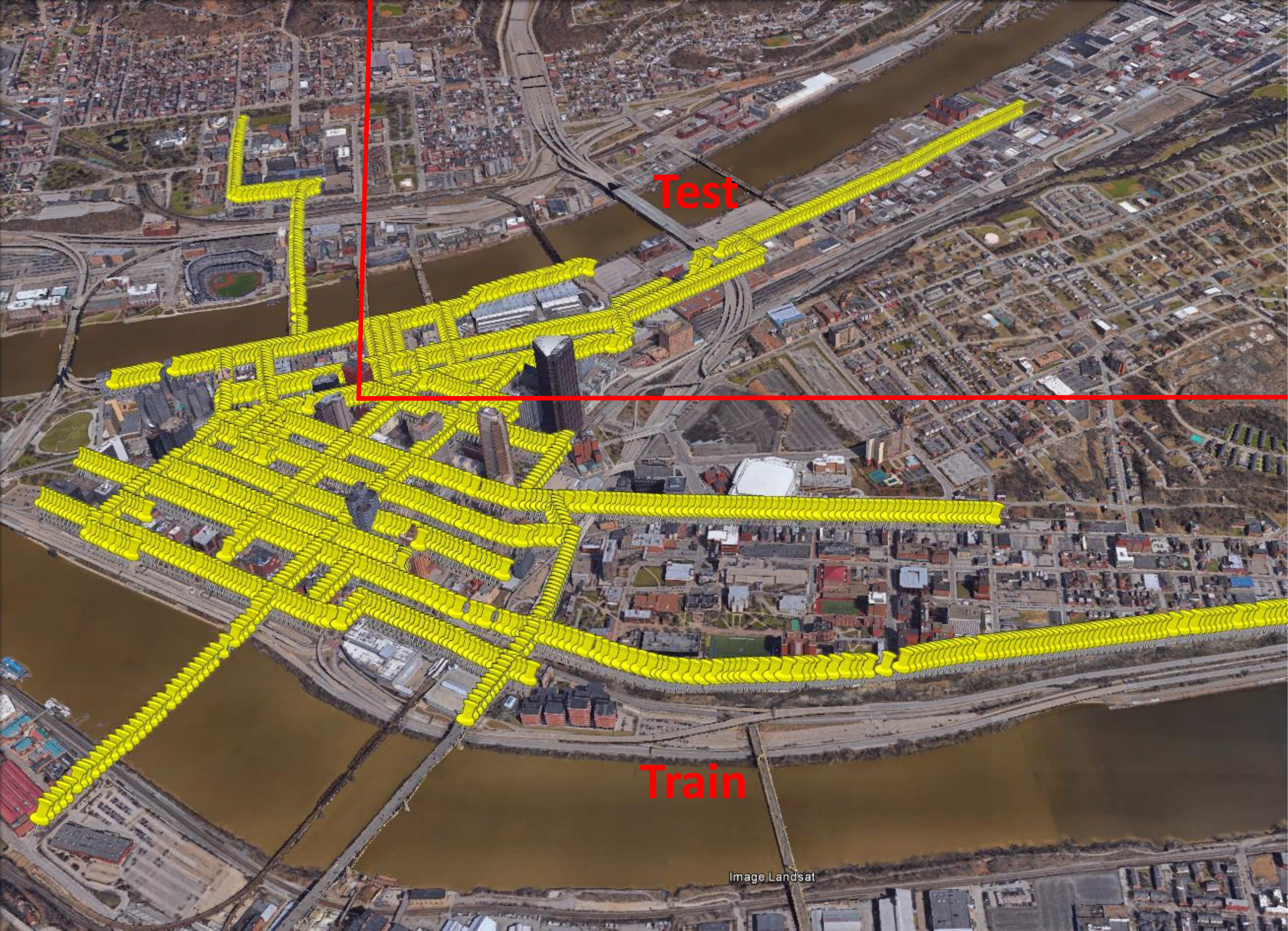}}
\subfigure{
\includegraphics[width=0.3\linewidth]{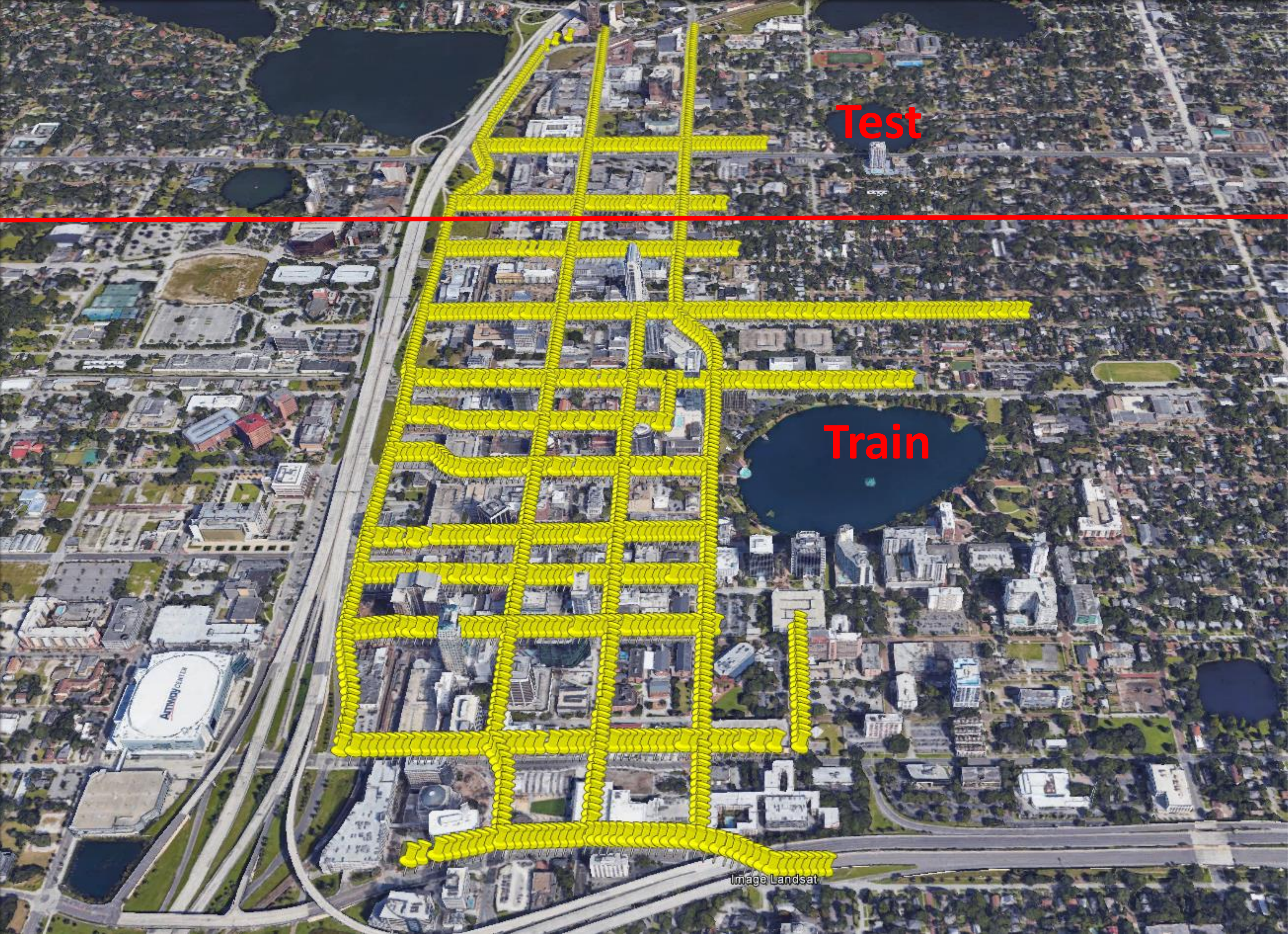}}
\subfigure{
\includegraphics[width=0.3\linewidth]{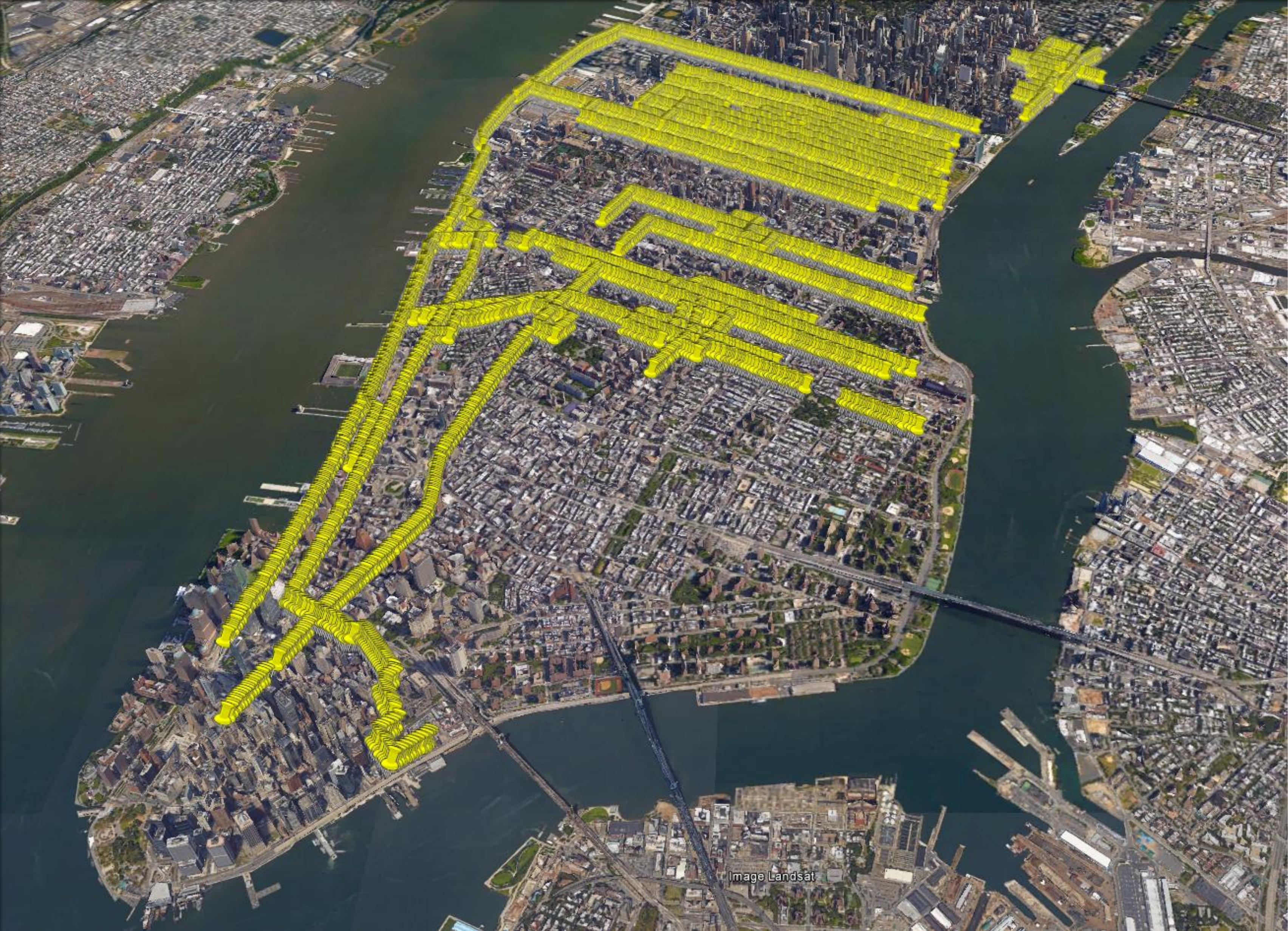}}
\caption{Sampled GPS locations in Pittsburg, Orlando and part of Manhattan.}
\label{fig:GPS}
\end{figure*}

To explore the geo-localization task using cross-view image matching, we have collected a new dataset of street view and bird's eye view image pairs around downtown Pittsburg, Orlando and part of Manhattan.  For this dataset we  use the list of GPS coordinates from Google Street View Dataset \cite{zamir2014}. The sampled GPS locations in the three cities are shown in Figure~ \ref{fig:GPS}. There are $1,586$, $1,324$ and $5,941$ GPS locations in Pittsburg, Orlando and Manhattan, respectively. We utilize DualMaps \footnote{http://www.mapchannels.com/DualMaps.aspx} to generate side-by-side street view and bird's eye view images at each GPS location with the same heading direction. The street view images are from Google and the overhead $45^{\circ}$ bird's eye view images are from Bing. For each GPS location, four image pairs are generated with camera heading directions of $0^\circ$, $90^\circ$, $180^\circ$ and $270^\circ$. In order to learn the deep network for building matching, we annotate corresponding buildings in every street view and bird's eye view image pair, which took roughly 300 hour of work.

Previous work on geo-localization by cross-view image matching have proposed several datasets. However, they are not suitable for our task. In the datasets presented in \cite{workman2015} and \cite{hays2016eccv}, a large portion of the images do not contain any building.
 Lin \etal\cite{lin2015} focus on matching cross-view buildings. However, the images in their collected dataset are aligned such that each image contains exactly one building. We explore geo-localization problem in urban environments by matching cross-view buildings. In our dataset, no careful image alignment is applied and every image usually contains multiple buildings.

\subsection{Experiments Setup}

To evaluate how the proposed approach generalizes to unseen city, we hold out all images from Manhattan exclusively for testing. Part of images from Pittsburg and Orlando are used for training. Since the sampled GPS locations are dense, one building may appear in multiple images with similar GPS coordinates. Especially, the bird's eye view images cover a relatively large area and may overlap with each other. Therefore, we divide images from Pittsburg and Orlando into training and test set based on the GPS coordinates. We take approximately one fifth of the images as training set and the rest as test set. The train-test split is shown in Figure~\ref{fig:GPS}.

In order to train building detectors, we annotate all buildings in around $7,000$ image pairs from training set. This results in $15k$ annotated buildings in street view and $40k$ annotated buildings in bird's eye view. A separate building detector is trained for street view and bird's eye view. We note that the building detectors generate high-accuracy results without the need to annotate buildings in the whole training set.

To learn the Siamese network, we annotate corresponding buildings in all the street view and bird's eye view image pairs from the training set. One Siamese network is learned by combining training data in Pittsburg and Orlando. Positive building pairs come from annotation and negative building pairs are randomly generated by pairing unmatched buildings. $15.7k$ positive building pairs are annotated for training. For both training and test sets, the number of negative building pairs is $20$ times more than that of positive building pairs. The geo-localization experiments are performed on a mixed test set of Pittsburgh and Orlando.

\subsection{Implementation Details}

To train the building detectors, the default setup of Faster R-CNN \cite{ren2015} is employed. Two building detectors are learned for street view images and bird's eye view images respectively.

For the Siamese network, the two sub-networks share the same architecture and weights. AlexNet \cite{krizhevsky2012imagenet} is used for the sub-networks. The learning rate of the last fully connected layer is set to $0.1$ and the learning rates of all the other layers are set to $0.001$. We use batch size of $128$. The image features obtained by the two sub-networks are fed into an L2 normalization layer separately before they are used to compute contrastive loss. The L2 normalization layer normalizes the two feature vectors to the same scale and make the network easier to learn. The Euclidean distance between two feature vectors is thus upper-bounded by $2$. The margin in the contrastive loss is set to $1$. We use the CNN trained on ImageNet \cite{krizhevsky2012imagenet} as pre-trained model and fine-tune it on our dataset.

For dominant sets, $\sigma$ is set to $0.3$ and $\alpha$ is set to $0.5$ when defining edge weights in graph $G$.

\subsection{Analysis of the Proposed Method}

\begin{figure}[htbp]
\begin{center}
   \includegraphics[width=0.9\linewidth]{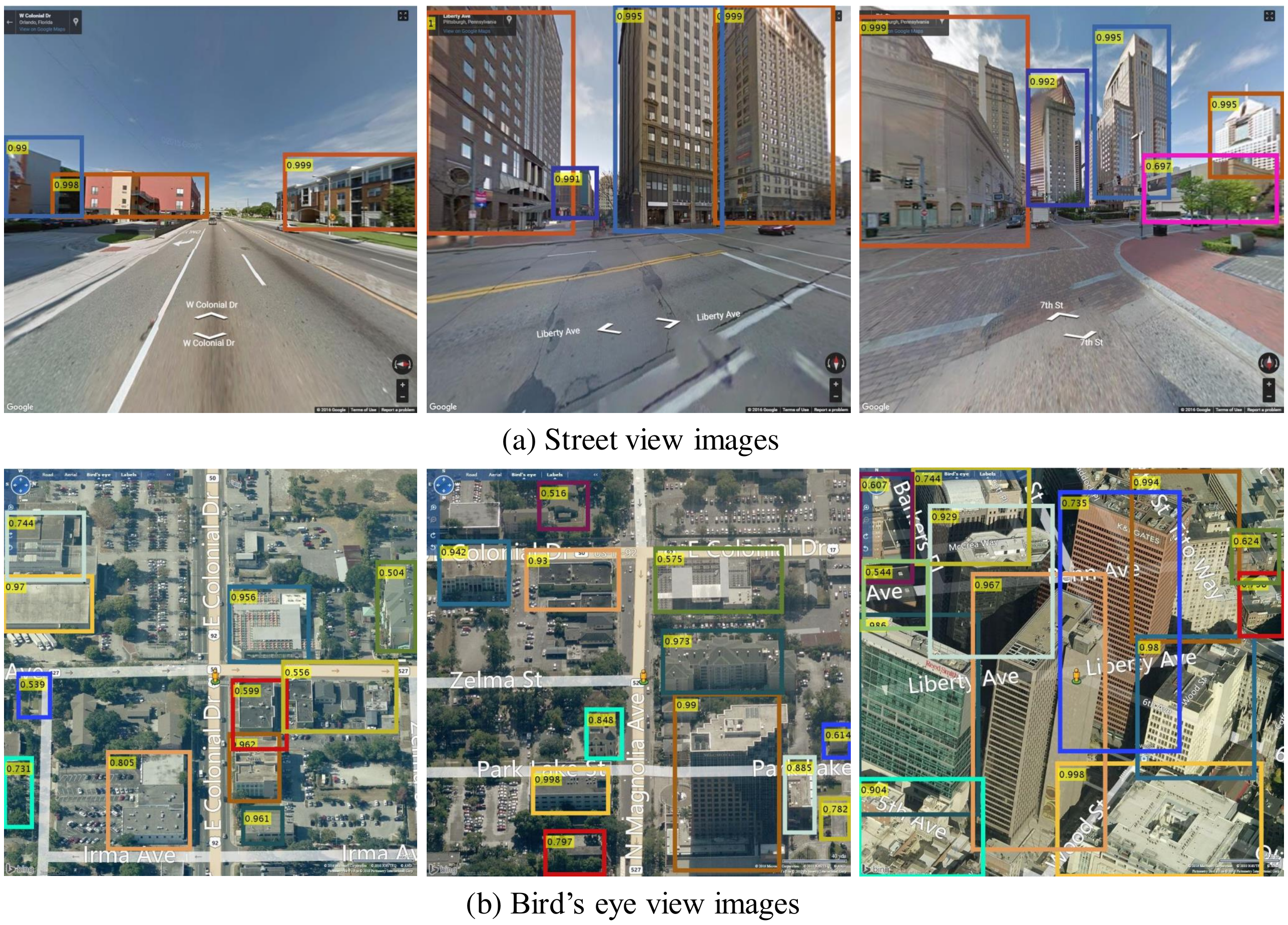}
\end{center}
   \caption{Building detection examples using Faster R-CNN.}
\label{fig:building_detection}
\end{figure}

\textbf{Building detection.} Figure~\ref{fig:building_detection} shows  examples of the building detection results in both street view and bird's eye view images. Each detected bounding box is assigned  a score. As evident from the figure, Faster R-CNN can achieve very good building detection results for both street view and bird's eye view images. Even for crowded scene where buildings occlude each other, Faster R-CNN is able to detect them successfully.

\begin{figure}[htbp]
\begin{center}
   \includegraphics[width=0.8\linewidth]{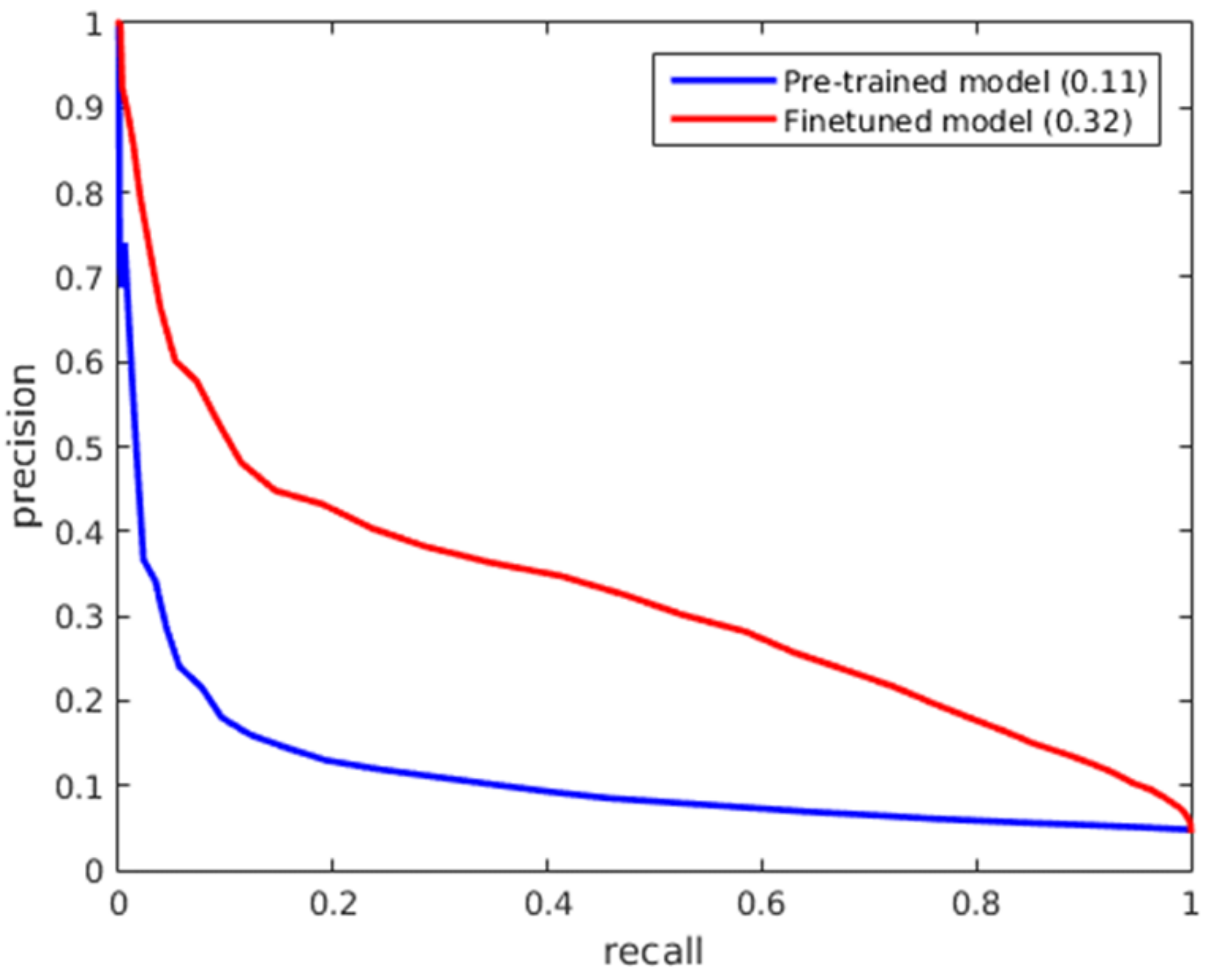}
\end{center}
   \caption{Precision-recall curves on test image pairs for cross-view building matching
 using pre-trained and fine-tuned models, respectively.}
\label{fig:building_match}
\end{figure}

\begin{figure}[htbp]
\begin{center}
   \includegraphics[width=0.95\linewidth]{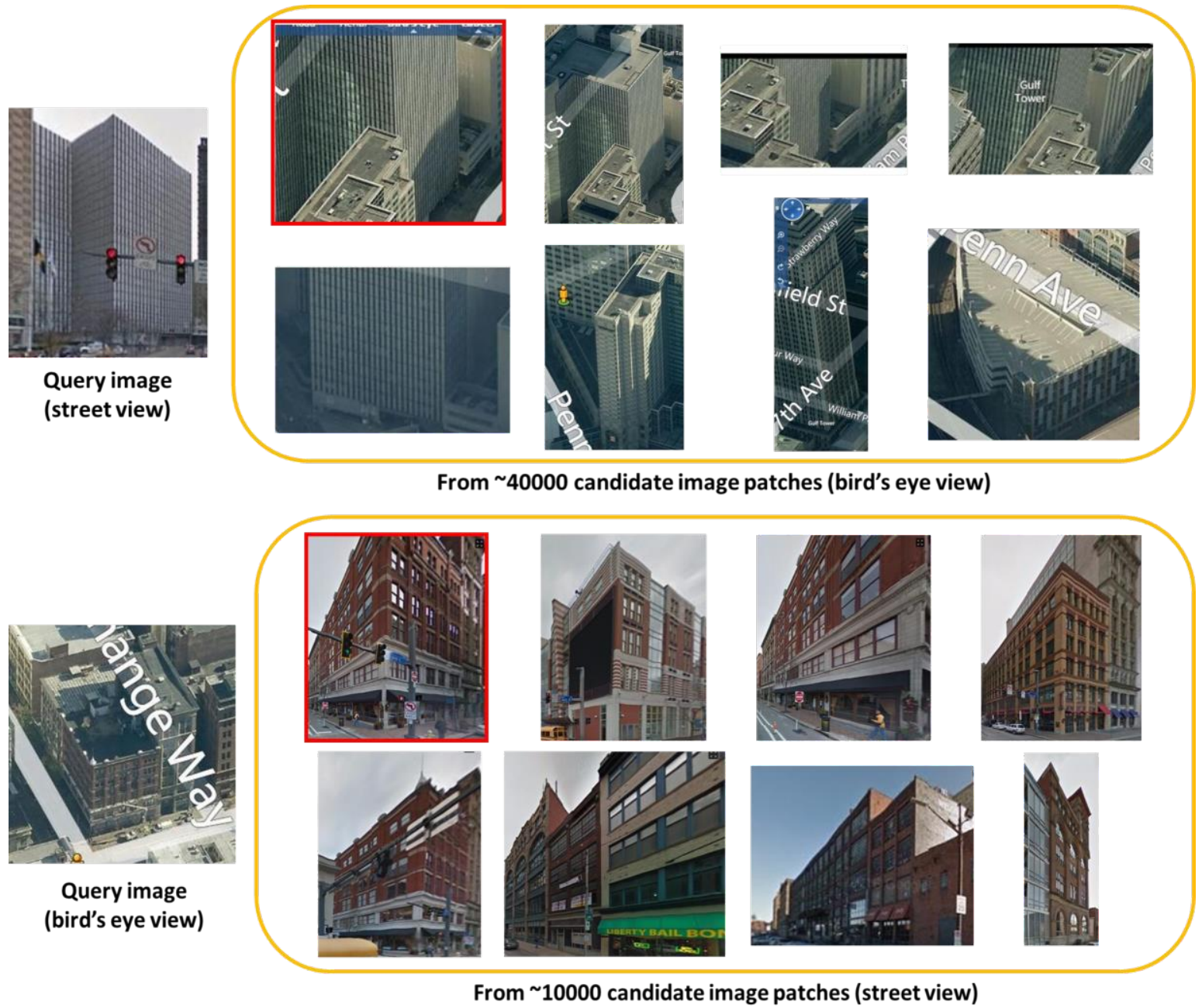}
\end{center}
   \caption{Visual examples of cross-view building matching results by our method. Red box indicates the correct match.}
\label{fig:building_match_visual}
\end{figure}

\textbf{Building matching.} To evaluate the building matching performance, we show the Precision-recall curves on test image pairs in Figure~\ref{fig:building_match}. Our fine-tuned model achieves average precision (AP) of $0.32$ compared that of $0.11$ for the pre-trained model. We also present visual examples of cross-view image matching in Figure~\ref{fig:building_match_visual}. The top 8 matched reference images are shown in the ranking order for each query image.

\begin{figure}[htbp]
	\centering
	\subfigure[]{
		\includegraphics[width=0.48\linewidth]{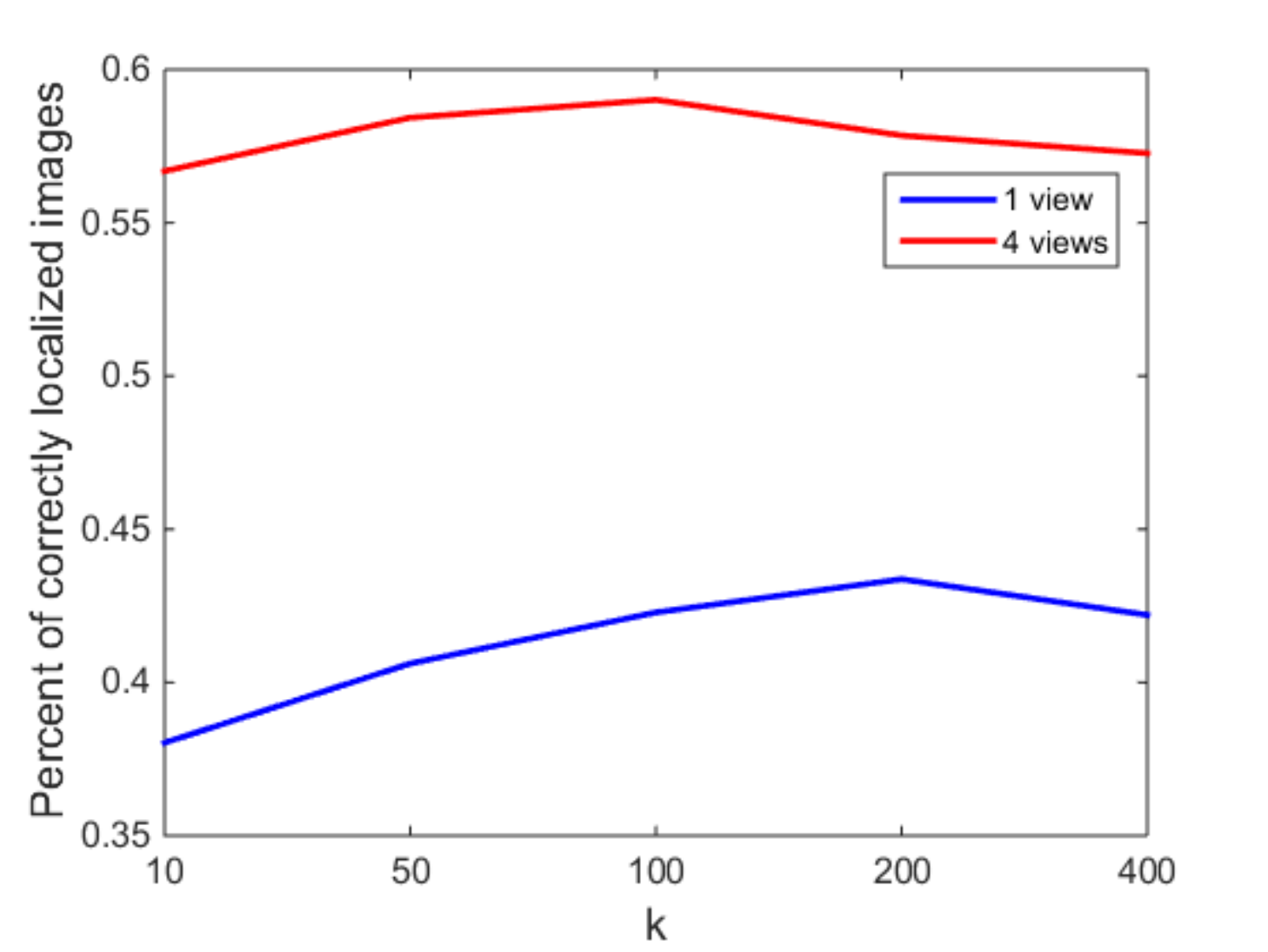}}
	\subfigure[]{
		\includegraphics[width=0.48\linewidth]{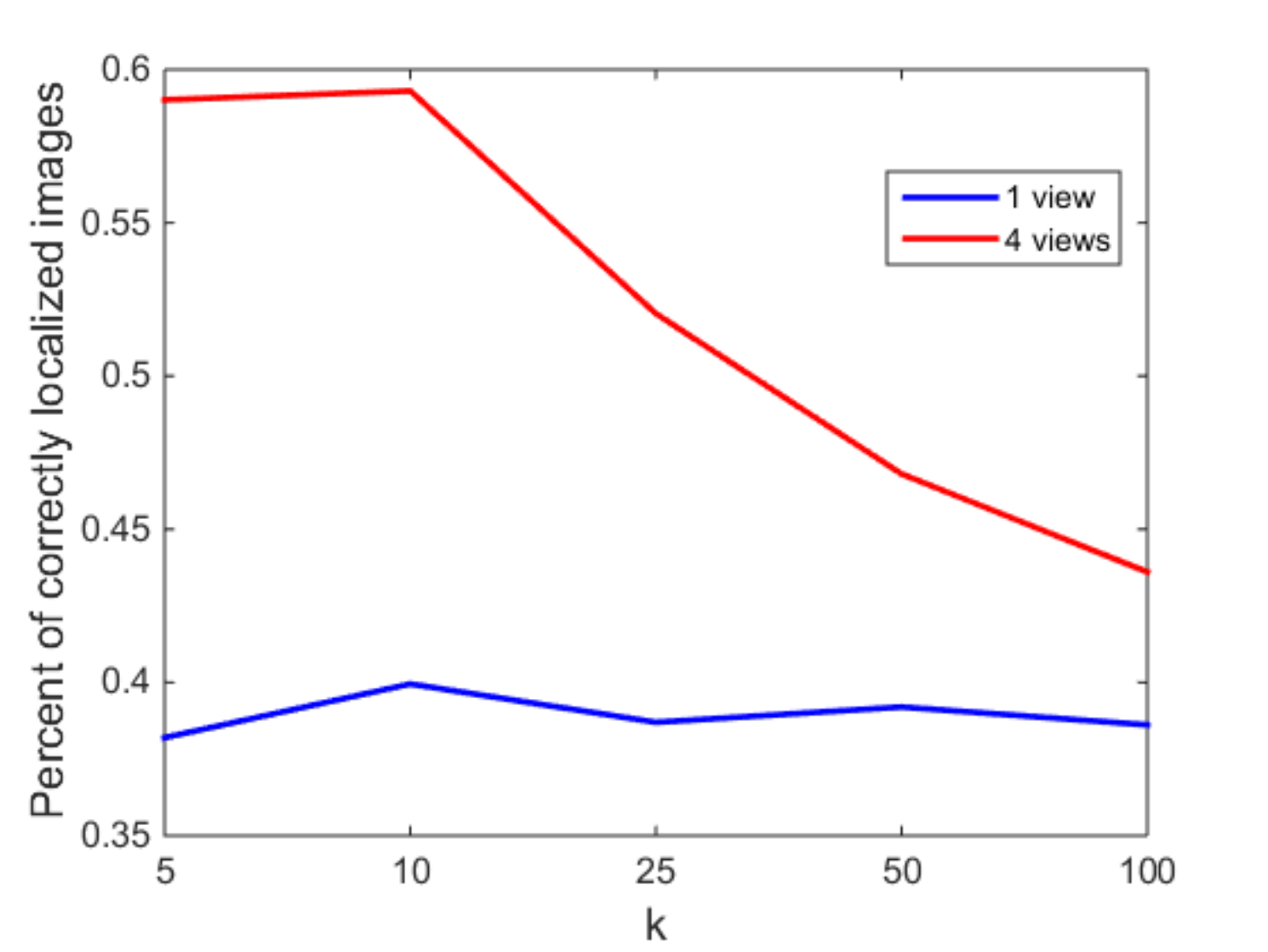}}
	\caption{Geo-localization results with different $k$ values. The error threshold is fixed as $300$m. (a) Results of using street view images as query and bird's eye view images as reference. (b) Results of using bird's eye view images as query and street view images as reference.}
	\label{fig:compareK}
\end{figure}

\textbf{Number of selected nearest reference neighbors ($k$).}
We compare the geo-localization result by varying the number of selected nearest reference neighbors, $k$ in Figure \ref{fig:compareK}. Street view images usually contain less buildings compared to bird's eye view images. Therefore, in order to achieve reasonable geo-localization results, more reference nearest neighbors should be considered when the query image is from street view. In our experiments, $k$ is set to $100$ when the query image is from street view while $k$ is set to $10$ when the query image is from bird's eye view.

\subsection{Comparison of the Geo-localization Results}
\begin{figure}[htbp]
	\centering
	
	\includegraphics[width=0.9\linewidth]{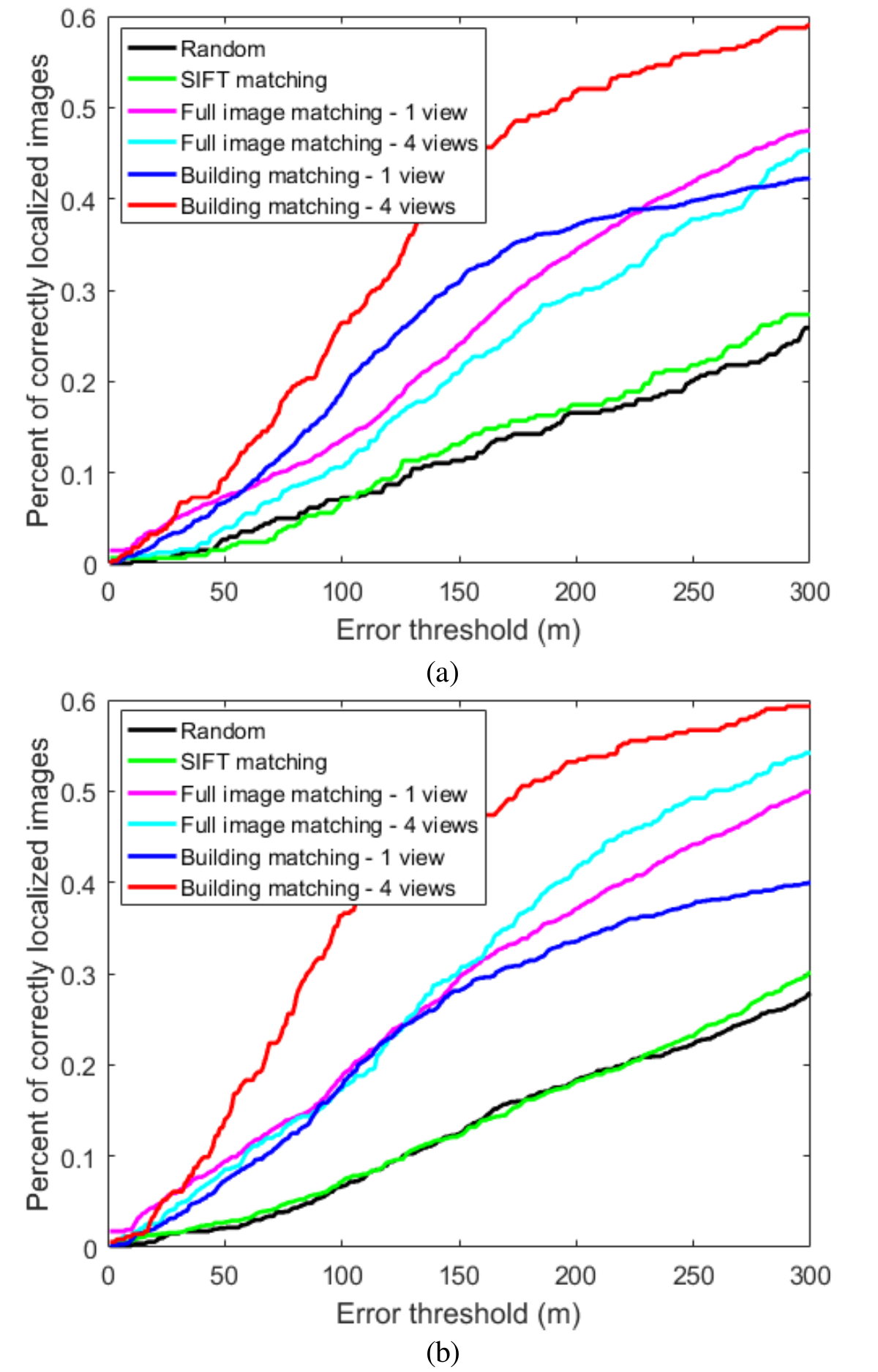}

	\caption{Geo-localization results with different error thresholds. (a) Results of using street view images as query and bird's eye view images as reference. (b) Results of using bird's eye view images as query and street view images as reference.}
	\label{fig:geolocalization_results}
\end{figure}

Figure~\ref{fig:geolocalization_results} compares the geo-localization results by using SIFT matching, random selection, building matching employing $1$ view query image and building matching using $4$ views query images (as shown in (Figure~\ref{fig:4views})).
For the approach of random image selection, we take the GPS of a randomly selected reference image as the final result for each query image.
It is obvious that geo-localization by building matching, which leverages the power of deep learning, outperforms that by matching hand-crafted local feature \ie SIFT. Also, our proposed approach outperforms random selection by a large margin. Moreover, query with $4$ images of four directions at one location improves the geo-localization accuracy by a large margin compared to using only $1$ image as a query.


\textbf{Building matching vs. full image matching.} To demonstrate the advantage of using building matching for cross-view image geo-localization, we conduct an experiment by training a Siamese network to match full images directly, which was used in the existing methods such as \cite{lin2015,hays2016eccv,workman2015}. No building detection is applied to images. Pairs of images taken at the same GPS location with the same camera heading direction are used as positive training pairs to Siamese network. Negative training image pairs are randomly sampled. The network structure and setup is the same as the Siamese network for building matching. During testing, the GPS location of a query image is determined by its best match and no multiple nearest neighbors matching process is necessary. Experiments using $1$ image as query and $4$ views as query images are performed and the results are illustrated in Figure \ref{fig:geolocalization_results}. Geo-localization by full image matching performs worse compared to building matching using $4$ views query images.

\begin{figure}[htbp]
\begin{center}
   \includegraphics[width=0.75\linewidth]{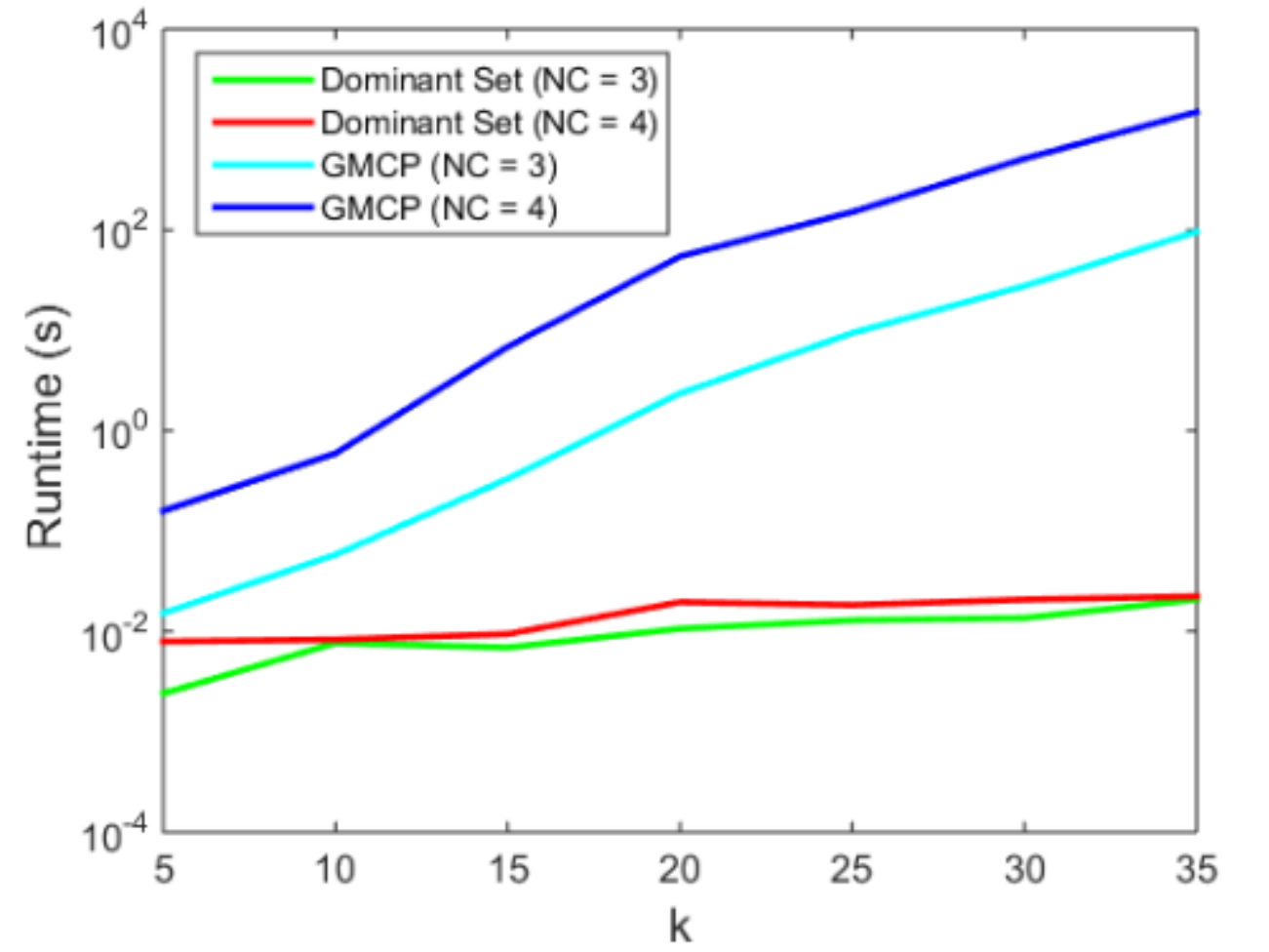}
\end{center}
   \caption{Runtime comparison of using dominant set and GMCP for multiple nearest neighbors matching.}
\label{fig:runtime}
\end{figure}

\begin{figure}[htbp]
\begin{center}
   \includegraphics[width=0.75\linewidth]{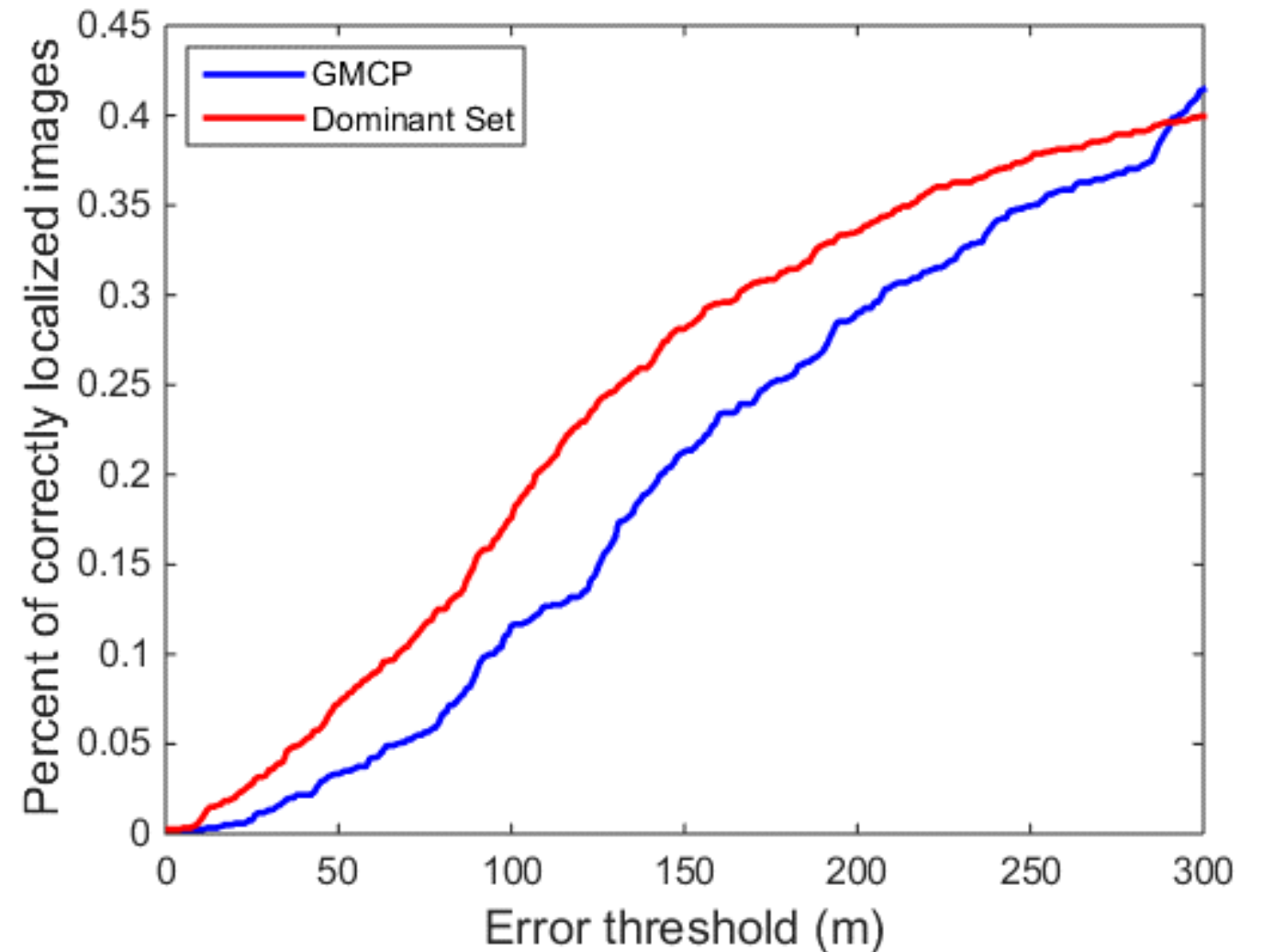}
\end{center}
   \caption{Geo-localization results comparison by using dominant set and GMCP. The experiment uses only $1$ view of bird's eye view image as query and $k$ is set to $10$.}
\label{fig:GMCPvsDS}
\end{figure}

\textbf{Dominant sets vs. GMCP \cite{feremans2003}.} To demonstrate the efficiency and effectiveness of using dominant sets for multiple nearest neighbors matching, we compare it with GMCP in terms of both runtime and performance. The runtime comparison is illustrated in Figure~\ref{fig:runtime}. The runtime of GMCP increases intensively by increasing either the number of clusters $NC$ or the number of nearest neighbors $k$. While dominant set is very efficient. Furthermore, we compare the geo-localization results by using dominant set and GMCP in Figure \ref{fig:GMCPvsDS}. Since the computational complexity of GMCP increases extremely fast when $NC$ or $k$ increases, and using GMCP to solve our problem is infeasible when $NC$ or $k$ is large, we conduct the experiment using $1$ bird's eye view image as query and set $k$ to $10$. For almost all the error thresholds, dominant set achieves better geo-localization accuracies than GMCP. In summary, using dominant set for multiple nearest neighbors matching in our geo-localization framework gives more accurate geo-localization results while being computationally efficient.

\subsection{Evaluation on Unseen Locations}
In this section, we verify if the proposed method can generalize to unseen cities. Specifically, we use images from the city of Pittsburgh and Orlando to train the model (building detection and building matching) and test it on images of the Manhattan area in New York city.

\begin{figure}[htbp]
	\centering
	\includegraphics[width=1.0\linewidth]{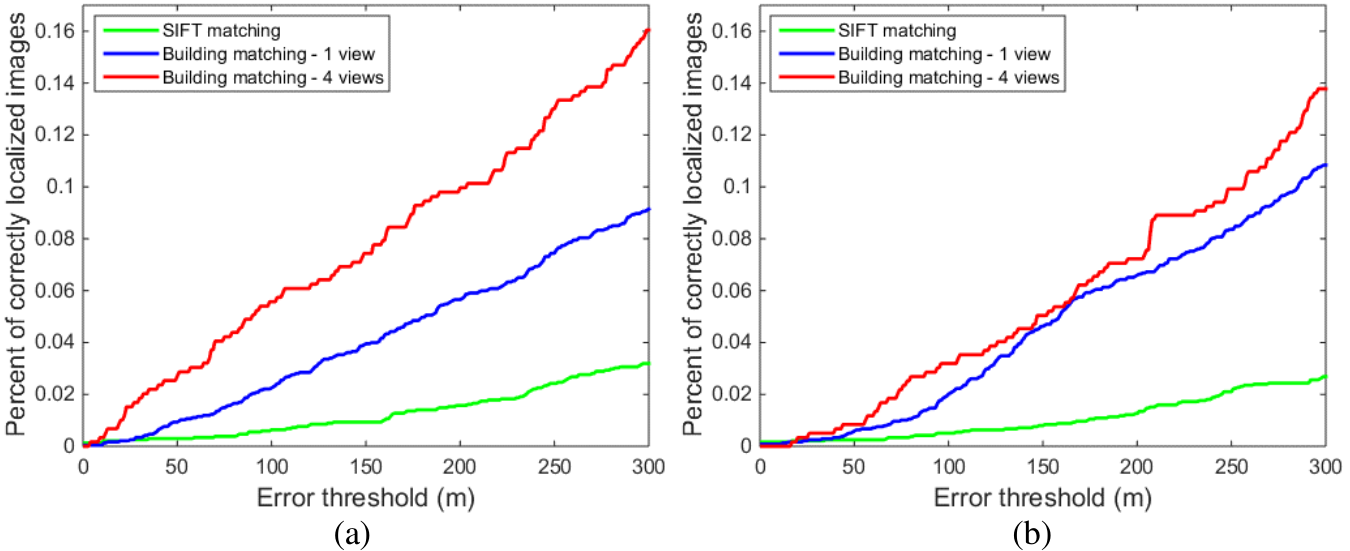}
	\caption{Geo-localization results on Manhattan images with different error thresholds. (a) Results of using street view images as query and bird's eye view images as reference. (b) Results of using bird's eye view images as query and street view images as reference.}
	\label{fig:Manhattan_results}
\end{figure}

As can be seen by the GPS locations in Manhattan area in Figure \ref{fig:GPS}, this geo-localization experiment works on city scale. In addition, tall and crowded buildings are common in Manhattan images, making the geo-localization task very challenging. The geo-localization results in the Manhattan area are shown in Figure \ref{fig:Manhattan_results}. The curves for Manhattan images are lower than those in Figure \ref{fig:geolocalization_results} because the test area in this experiment is much larger. The fact that our geo-localization results are still much better than the baseline method - SIFT matching demonstrate the ability of generalization of our proposed approach to unseen cities.


\section{Conclusion}
\label{sec:conclusion}

In this paper we propose an effective framework of cross-view image matching for geo-localization, which localizes a query image by matching it to a database of geo-tagged images in the other view. Our approach utilizes deep learning based techniques for building detection and cross-view building matching. The final geo-localization results are achieved by matching multiple query buildings using dominant sets. In addition, we introduce a new large scale cross-view dataset consisting of pairs of street view and bird's eye view images. On this dataset, the experiments show that our method outperforms other approaches for cross-view geo-localization. In our future work, we are going to extend our approach to areas that may not contain any building by exploring matching other objects and semantic information, \eg road structure, water reservoirs, etc. In that case, the idea of buildings matching can be generalized to multiple attributes matching.

{\small
\bibliographystyle{ieee}

}

\end{document}